\documentclass[runningheads]{llncs}
\usepackage{arydshln}
\usepackage{xcolor}
\usepackage[normalem]{ulem}
\usepackage{booktabs}
\usepackage{subcaption}
\usepackage{fontspec}
\newfontfamily\arabicfont[ Script=Arabic]{Amiri-Regular.ttf}

\newcommand{\ar}[1]{{\arabicfont #1}}
\usepackage{makecell}   

\usepackage[T1]{fontenc}
\usepackage{multirow}

\usepackage{graphicx}
\usepackage{amsmath} 
\begin{document}
\hyphenation{manu-script}
%

\title{Cross-Lingual Learning within Arabic Script for Low-Resource HTR}


%
%

\author{Sana Al-azzawi\orcidID{0000-0001-7924-4953} \and
Elisa Barney\orcidID{0000-0003-2039-3844} \and
Marcus Liwicki\orcidID{0000-0003-4029-6574}}
\authorrunning{S. Al-azzawi et al.}

\institute{
Luleå University of Technology\\
Department of Computer Science, Electrical and Space Engineering\\
Luleå, 97187, Sweden\\
\email{sana.al-azzawi@ltu.se, elisa.barney@ltu.se, marcus.liwicki@ltu.se}
}
\date{}
\maketitle  





\begin{abstract}
Handwritten Text Recognition (HTR) with limited labeled data remains a challenging problem, particularly for Arabic-script languages. Although modern sequence-based recognizers perform well in high-resource settings,
their accuracy degrades sharply as training data becomes scarce. Arabic-script languages share a common writing system with substantial character overlap, motivating cross-lingual learning as a strategy to mitigate data scarcity.
We conduct a controlled line-level study of cross-lingual joint training for Arabic-script HTR under low-resource regimes (number of samples $K \in {100, 500, 1000}$ labeled lines) on Arabic (KHATT), Urdu (NUST-UHWR) and Persian (PHTD). CRNN and Vision Transformer-based HTR-VT models are trained on the union of multiple related Arabic-script datasets to mitigate the data scarcity and are evaluated on individual target languages. Both architectures benefit from cross-language training under low-resource conditions. CRNN remains more effective under extremely limited target-language data, whereas the benefits of cross-language training for HTR-VT become less consistent as larger amounts of target-language data become available.
On Persian (PHTD), joint training achieves a Character Error Rate (CER) of 9.99 \%, surpassing previously reported results despite not using the full available training data. On an additional Urdu dataset (UNHD), joint training reduces CER from 17.20\% to 14.45\%. 





\end{abstract}
\section{Introduction}
Handwritten text recognition (HTR) for Arabic-script languages remains challenging, particularly under low-resource conditions where only limited labeled data is available. While modern neural recognizers achieve strong performance when trained on large-scale datasets, their accuracy degrades substantially when labeled training data is scarce \cite{luo2020learn}.
This challenge is especially pronounced for Arabic-script HTR due to its cursive connectivity, position-dependent letter forms, and the high visual similarity among many characters that differ only by dots or diacritics \cite{salaheldin2025advancements}.

In contrast to Latin-script HTR, where large standardized datasets exist across multiple languages, Arabic-script HTR resources remain comparatively limited and fragmented, particularly for multilingual scenarios, which are mostly available at the line level. Many Arabic-script datasets focus on character- or word-level recognition, with fewer large-scale line-level resources available. Publicly available page-level datasets are currently available primarily for Arabic, while Urdu and Persian resources remain more limited. For this reason, we focus on line-level recognition in this work.


Furthermore, Arabic script datasets are imbalanced in terms of scale (some are just a few hundred pages, some just a few hundred lines), raising concerns about the feasibility of training robust HTR systems under such constraints.
In practice, while labeled data in the target language remains scarce, additional handwritten material from related scripts can already be accessible through public digitized archives or web-collected document repositories.
This motivates the investigation in this paper: when only a limited number of annotated pages or labeled lines are available for a target language, \textbf{can data from related Arabic-script languages be leveraged to improve recognition under low-resource conditions? } 
\newline
\newline
\noindent\textbf{Contributions.}
This paper makes the following contributions:
\begin{itemize}
    \item We present a controlled line-level joint training study across three distinct Arabic-script languages (Arabic, Urdu, and Persian) under strictly low-resource regimes ($K \in \{100, 500, 1000\}$ labeled lines), comparing both CRNN- and Vision Transformer-based HTR-VT architectures.
    

    \item  We demonstrate that cross-lingual joint training consistently improves recognition accuracy when the target dataset is limited in size and textual diversity across Arabic, Urdu, and Persian datasets.

    \item We provide a detailed character-level analysis showing that transfer gains are structured: improvements concentrate on characters shared across scripts, while language-specific characters remain largely stable.
    

    \item We provide a dataset-level statistical analysis of textual diversity and duplication, showing how these factors influence cross-lingual transfer.
\item We show that CRNN models remain more data-efficient under low-resource conditions and consistently benefit from cross-language training, whereas HTR-VT benefits mainly under limited target-language data and sometimes exhibits degraded performance as more target-language data becomes available.
    
\end{itemize}
All code, splits, training configurations, and evaluation protocols are released to ensure full reproducibility.\footnote{GitHub link will be publicly released upon acceptance}

The remainder of this paper is organized as follows. Section~\ref{sec:background} summarizes structural overlap in Arabic-script languages.  Section~\ref{sec:related} reviews related work. Section~\ref{sec:methodology} describes the problem formulation, training paradigms, and recognition architectures. Section~\ref{sec:experiment} details the datasets, implementation settings, and evaluation protocol. Section~\ref{sec:results} presents quantitative results and character-level analyses. Finally, Section~\ref{sec:conclusion} concludes the paper and outlines future directions.
\section{Background: Structural Overlap in Arabic Scripts} \label{sec:background}

\begin{figure}
\centering
\includegraphics[width=0.55\textwidth]{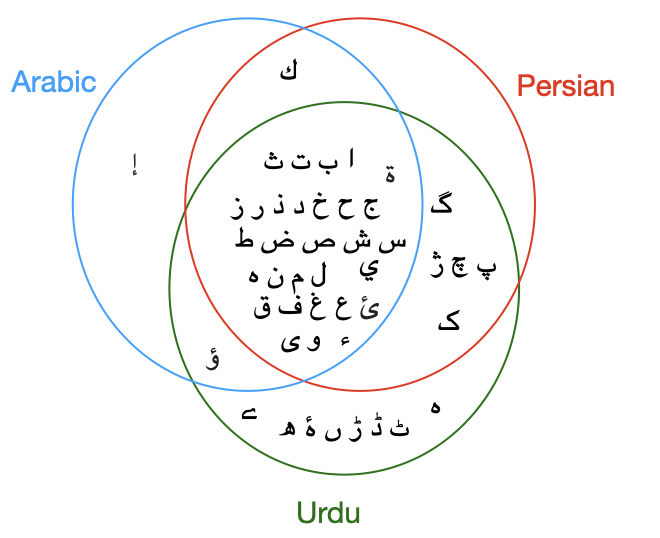}
\caption{Overlap between the character inventories of Arabic, Persian, and Urdu derived from the datasets used in this study. The diagram shows shared and language-specific (unique) characters in the experimental setting.} 
\label{fig:venn}
\end{figure}

Languages that use the Arabic-script share a largely common writing system based on a common set of base letterforms that are often distinguished only by dots or diacritic marks~\cite{ahmad2017impact}. As shown in Fig.~\ref{fig:venn}, the character inventories of Arabic, Persian and Urdu overlap extensively: 30 characters are shared across all three scripts, while Urdu introduces nine additional characters, Persian introduces none, and Arabic contains a single unique character. 
Seven are shared across pairs of languages. Beyond character-level similarity, these languages are historically interconnected, with substantial lexical borrowing and recurring character sequences across corpora \cite{salaheldin2025advancements,nemati2022persian}. This structural and statistical overlap suggests that cross-lingual transfer may be particularly effective within the Arabic writing system.

\section{Related Work}\label{sec:related}

Prior work in multilingual printed OCR shows that LSTM-based models trained on mixtures of Latin-script languages can generalize across languages within the same script family~\cite{ul2013can}. More recently, task grouping strategies for multilingual OCR demonstrated that automatically grouping structurally related languages outperforms fully shared or fully separated architectures~\cite{huang2022task}.


Within Arabic-script recognition, prior work has explored transfer in several related but distinct directions.  Cross-script learning for synthetic and printed Arabic-like scripts has been explored using MDLSTM architectures~\cite{ahmad2017impact}, showing that recognition benefits when scripts share visual similarity. However, these experiments were conducted on printed and synthetic data rather than real handwritten documents.
Low-resource printed Ottoman recognition has been addressed through synthetic data generation and transfer learning from Arabic~\cite{bilgin2023printed}. Although additional data mitigates scarcity, cross-language transfer between Arabic and Ottoman leads to performance degradation, and the analysis remains limited to printed OCR rather than handwritten HTR.


Transfer between modern Arabic handwriting and historical Ottoman manu\-scripts has been studied using transformer-based models~\cite{broadwell2025multilingual}. These experiments consider a single auxiliary–target pair and do not isolate the effects of script relatedness under strictly low-resource supervision.

Several works combine printed and handwritten Urdu datasets to improve handwritten recognition performance~\cite{riaz2022conv,hamza2024network}. These approaches increase training data volume within the same language but do not examine cross-lingual transfer across distinct Arabic-script languages.

Bilingual training across disjoint scripts, such as Urdu and English, has also been explored to assess architectural robustness in higher-resource settings~\cite{ul2022convolutional}. However, this work does not systematically analyze low-resource cross-language behavior within the Arabic-script family.

Despite these efforts, prior studies focus on printed text, high-resource regimes, or single auxiliary–target pairs, often within the same language. A controlled line-level study across multiple closely related Arabic-script languages under strictly low-resource handwritten conditions remains largely unexplored. This motivates our study.

\section{Methodology }\label{sec:methodology}
We exploit the character overlap across Arabic, Urdu, and Persian to evaluate how cross-lingual learning improves recognition when target data is limited.

\subsection{Problem Setting}
We consider line-level HTR for Arabic-script languages under low-resource conditions. 
Let $\mathcal{D}_t$ denote a target-language dataset and let the auxiliary dataset $\mathcal{D}_a = \bigcup_{j=1}^{J} \mathcal{D}_a^{(j)}$, where $\mathcal{D}_a^{(j)}$ denotes the $j$-th language specific auxiliary dataset and 
$J$ is the number of auxiliary datasets used.
Each dataset consists of pairs $(x_i, y_i)$, where $x_i$ is a text-line image and $y_i$ its corresponding character-sequence transcription.
To evaluate low-resource scenarios, we restrict the labeled training subset of the target dataset $\mathcal{D}_t$ to $K \in \{100, 500, 1000\}$ randomly sampled line images. Auxiliary datasets are used with their complete training splits. Evaluation is performed independently for each target language on its full official test split using Character Error Rate (CER) as the metric.

\subsection{Recognition model}

We use two recognition architectures: a CNN-based CRNN model \cite{retsinas2022best} and the transformer-based HTR-VT model \cite{li2025htr}, both adapted for Arabic-script HTR.

The CRNN consists of a convolutional feature extractor with residual convolutional blocks followed by three stacked bidirectional LSTM layers and is trained using the CTC objective. Sequential features are obtained through column-wise pooling over convolutional feature maps.

HTR-VT combines a CNN feature extractor with a 4-layer Vision Transformer (ViT) encoder \cite{dosovitskiyimage} using 6 attention heads and predicts character sequences using CTC loss. Following \cite{li2025htr}, span masking, the Sharpness-Aware Minimization (SAM) optimizer, and exponential moving average (EMA) are used during training.

\subsection{Training Paradigms}
In our study, three Arabic-script languages are considered: Arabic, Urdu, and Persian, for which we use the datasets: KHATT, NUST-UHWR, and PHTD. Each language's dataset serves as the target dataset $\mathcal{D}_t$ in turn, while the remaining datasets are treated as auxiliary data $\mathcal{D}_a$. 
In this paper, we use the term joint training to denote training a single HTR model on the combined training data of multiple languages.
Experiments are conducted independently for each target language under identical low-resource conditions.

\begin{figure}[t]
\centering
\includegraphics[width=0.9\textwidth]{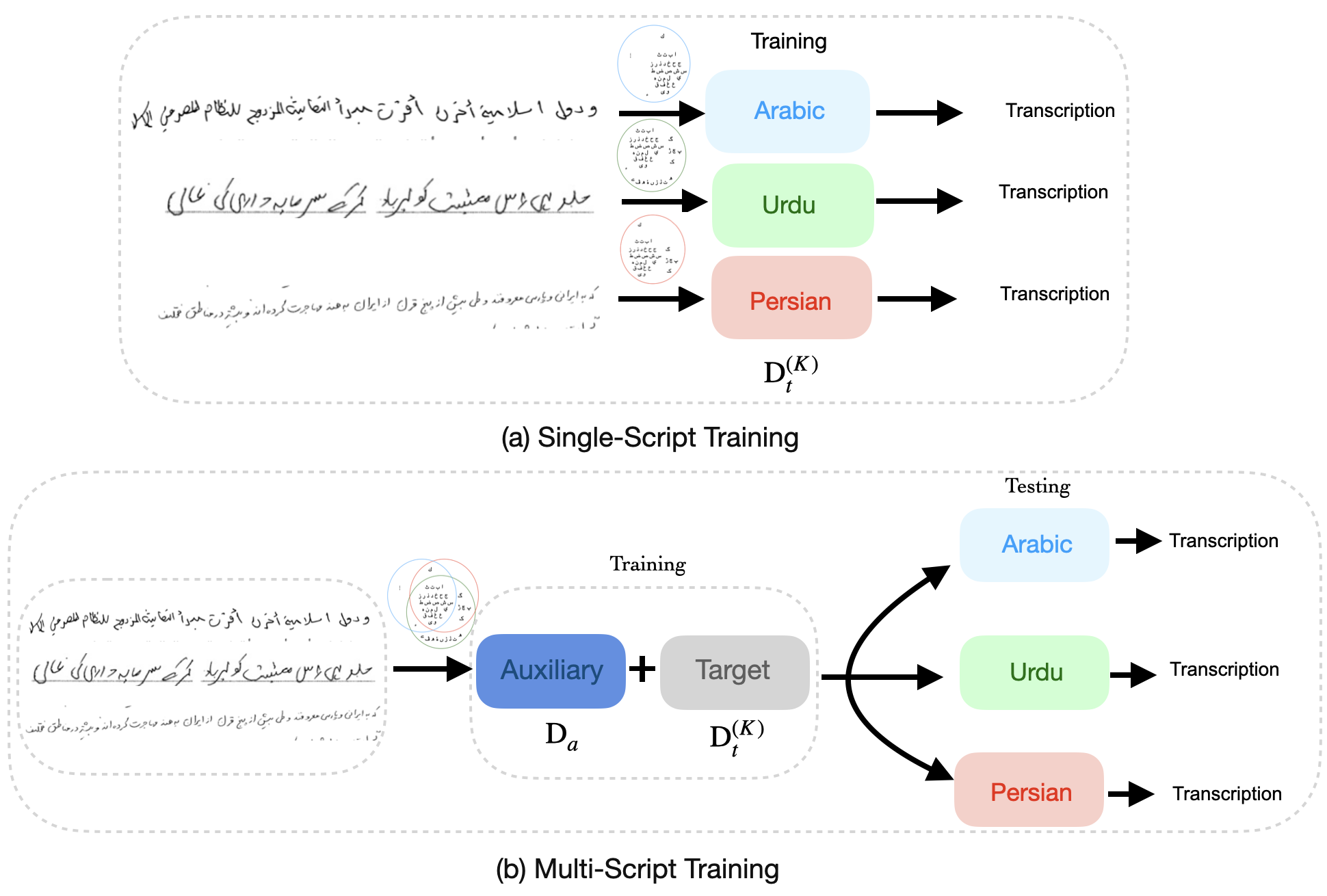}
\caption{Comparison of training paradigms. 
(a) Single-script ($J=0$): training and evaluation are performed using only the target dataset $\mathcal{D}_t^{(K)}$ with $K$ labeled samples. 
(b) Multi-script ($J=1$ or $J=2$): training is performed on the union of the limited target subset $\mathcal{D}_t^{(K)}$ and auxiliary datasets $\mathcal{D}_a^{(j)}$ (merged as $\mathcal{D}_a$), while evaluation is conducted on each target language separately.}

\label{fig:arch}
\end{figure}

\paragraph{Single-Script Training.}
In the single-script setting (Fig.~\ref{fig:arch}a), a separate recognizer is trained for each target language using only data from its corresponding dataset. The model is trained and evaluated exclusively on a single script (Arabic, Urdu, or Persian), without exposure to any auxiliary languages (i.e., $J = 0$). To simulate low-resource conditions, we restrict the labeled training subset of the target dataset $\mathcal{D}_t^{(K)}$ to $K \in \{100, 500, 1000\}$ randomly sampled line images.

\paragraph{Multi-Script Joint Training.}
In the multi-script setting (Fig.~\ref{fig:arch}b), the model is trained jointly on the union of the limited target subset $\mathcal{D}_t^{(K)}$ and the auxiliary data $\mathcal{D}_a = \bigcup_{j=1}^{J} \mathcal{D}_a^{(j)}$, where $J \geq 1$ from other languages. Formally,
\begin{equation}
\mathcal{D}_{train} = \mathcal{D}_t^{(K)} \cup \mathcal{D}_a .
\label{eq:train_union}
\end{equation}
Auxiliary datasets are used in full while the target remains limited to K labeled lines. This mirrors a practical case where related language datasets are available, but annotated target-language data is limited.

We consider two variants of multi-script training. In the \emph{Multi-script ($J=1$ aux.)} setting, the target dataset is combined with a single auxiliary dataset. In the \emph{Multi-script ($J=2$ aux.)} setting, the target dataset is combined with both remaining Arabic-script datasets. Both variants involve cross-lingual transfer; they differ only in the amount of auxiliary data included during training.

\section{Experiment}\label{sec:experiment}
This section describes the datasets, experimental protocol, and evaluation settings used to assess the proposed cross-lingual joint training strategy. 
\subsection{Datasets}
\begin{figure}[t]
\centering
\includegraphics[width=0.70\textwidth]{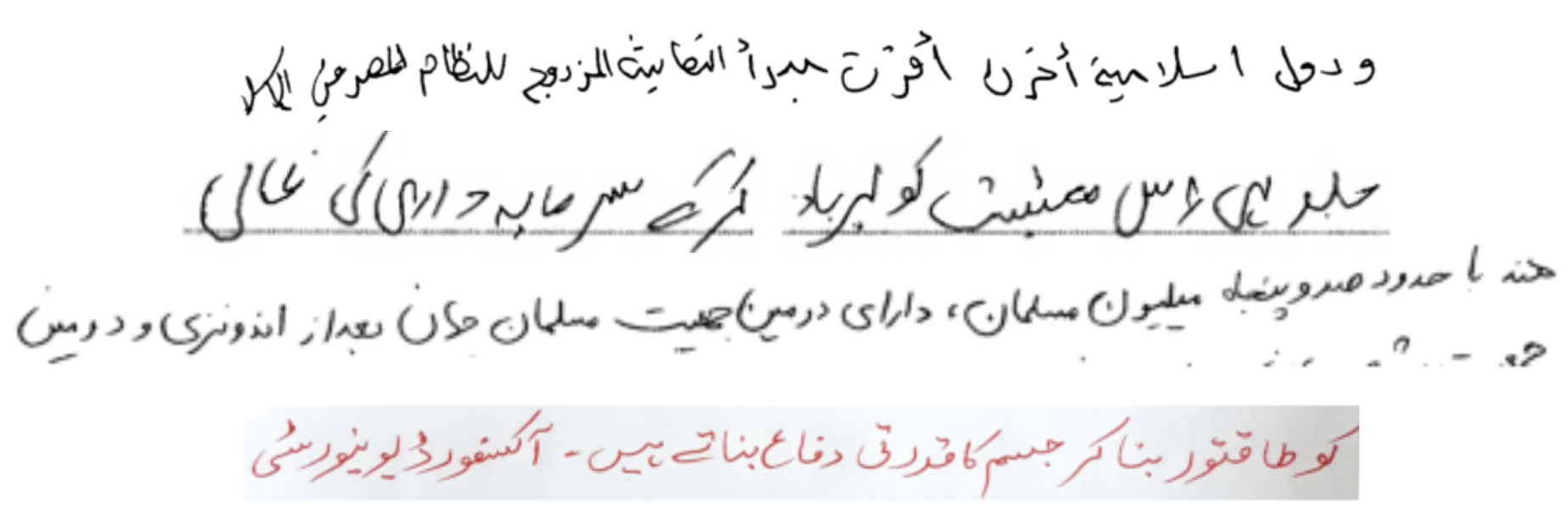}
\caption{ Examples of handwritten text lines from the four datasets used in this study. From top to bottom: KHATT (Arabic), NUST-UHWR (Urdu), PHTD (Persian), and UNHD (Urdu).} \label{fig:dataset}
\end{figure}

We use four publicly available Arabic-script HTR datasets in our experiments: KHATT (Arabic), NUST-UHWR (Urdu), PHTD (Persian), and UNHD (Urdu). The first three serve as the primary datasets for controlled cross-lingual evaluation, while UNHD is used as an additional low-resource benchmark to confirm our results. A representative example from each dataset is shown in Fig.~\ref{fig:dataset} and details are described next.


\paragraph{KHATT (Arabic).}
KHATT~\cite{mahmoud2014khatt} is a widely used benchmark for modern Arabic HTR. It consists of paragraph images written by 1,000 writers from 18 countries. Following common practice, we use the subset of 2,000 unique paragraphs, which are segmented into line images using the official split. The resulting partition contains 4,837 training lines, 938 validation lines, and 967 test lines.

\paragraph{NUST-UHWR (Urdu).}
NUST-UHWR~\cite{ul2022convolutional} is a large-scale unconstrained Urdu handwritten text dataset written in Nasta’liq style. It contains 10,605 text lines produced by nearly 1,000 contributors and spans multiple topical domains to ensure linguistic diversity. The official split includes 8,483 training lines, 1,061 validation lines, and 1,061 test lines. 

\paragraph{PHTD (Persian).}
The Persian Handwritten Text Dataset (PHTD)~\cite{alaei2012dataset} consists of 140 handwritten pages written by 47 writers. As the original release provides page-level images with pixel-level masks but no predefined line images or official splits, we use the line-level version and leakage-free page-based partition introduced in~\cite{al2026cer}. The resulting split contains 1,473 training lines, 160 validation lines, and 155 test lines. Due to its comparatively small size and limited textual diversity, PHTD is particularly suitable for evaluating transfer behavior under low-resource conditions.

\paragraph{UNHD (Urdu, Additional Low-Resource Dataset).}
UNHD (Urdu Nasta’liq Handwritten Dataset)~\cite{ahmed2019handwritten} contains 10,000 handwritten text lines collected from 500 writers. However, only approximately 600 training lines are unique in terms of textual content, resulting in limited effective lexical diversity. To evaluate the robustness of cross-lingual transfer in an additional low-resource setting, we use the publicly available unique subset.

\subsection{Training and Implementation Details}

All experiments were implemented in PyTorch and conducted on NVIDIA A100-SXM4-40GB GPUs. 

Line images were converted to grayscale and size-normalized using adaptive resizing while preserving the aspect ratio. 
For the CRNN model, images were adaptively resized to a target height of 110 pixels with a maximum width of 1450 pixels, where the target dimensions were derived from the average text-line dimensions across the primary datasets. For HTR-VT, we follow \cite{li2025htr} and resize images to a fixed input size of $512\times64$ while preserving aspect ratio. 
To account for the right-to-left nature of Arabic-script writing while maintaining left-to-right feature extraction, ground-truth transcriptions were reversed during training. A consistent augmentation pipeline was applied across datasets and training paradigms, including affine transformations, elastic distortion, morphological operations, brightness and contrast adjustment, and gamma correction.

Both models were trained using batch size 16. The CRNN model was optimized using AdamW with an initial learning rate of $5\times10^{-4}$ and a multi-step learning-rate schedule. HTR-VT was trained following \cite{li2025htr} using AdamW with SAM, EMA, cosine warmup scheduling, span masking and a maximum learning rate of $5\times10^{-4}$. For each $K$, all models were trained for a fixed number of optimization steps, and the checkpoint with the best validation CER was selected for evaluation.

\subsection{Experimental Protocol}

For each target language and each low-resource regime 
$K \in \{100, 500, 1000\}$, we evaluate the following setups:
\begin{enumerate}
    \item \textbf{Single-script training ($J=0$):} the model is trained using only 
    $K$ labeled lines from the target dataset.
    
    \item \textbf{Multi-script ($J=1$ aux.):} the model is trained using 
    $K$ target lines together with one auxiliary Arabic-script dataset.
    
    \item \textbf{Multi-script ($J=2$ aux.):} the model is trained using 
    $K$ target lines together with both remaining Arabic-script datasets.
\end{enumerate}

For each value of $K$, \textbf{ target subsets were sampled using fixed random seeds} to ensure consistent comparisons across single- and multi-script settings. A unified character vocabulary was constructed as the union of characters across all involved languages. Recognition performance is evaluated using CER, computed as the normalized Levenshtein distance between the predicted and ground-truth character sequences, where lower CER indicates higher transcription accuracy.

Across all configurations, the recognition architectures, optimization settings, augmentation strategy, and decoding procedure are kept consistent, and results are obtained using CTC greedy decoding without any external language model or lexicon.

\section{Results}\label{sec:results}

This section reports the evaluation of cross-lingual learning under low-resource conditions. We begin by analyzing single-script behavior, then examine the effect of one- and two-auxiliary joint training across Arabic, Urdu, and Persian datasets. We further evaluate generalization to an additional low-resource dataset and present a character-level analysis to interpret the transfer behavior.

Table~\ref{tab:single_partial_full_across_k} summarizes recognition performance under single-script and multi-script training across three low-resource regimes ($K \in \{100, 500, 1000\}$).
We also evaluated a balanced sampling strategy in which each batch
contained an equal number of target and auxiliary samples. This approach did not yield consistent improvements over proportional mixing; therefore, all reported results use the standard mixed training procedure.
Table~\ref{tab:fullCER} shows results when trained on the full target datasets for $J=0$ and $J=2$, as well as comparing to results from the literature.

\begin{table*}[t]
\centering
\caption{Cross-lingual training results under low-resource ($K{=}100, 500, 1000$). The table compares CRNN and HTR-VT under single-script ($J{=}0$), one-auxiliary ($J{=}1$), and two-auxiliary ($J{=}2$) settings. Best results for each dataset and $K$ are shown in bold.}
\label{tab:single_partial_full_across_k}
\small
\resizebox{0.98\textwidth}{!}{%
\begin{tabular}{clccccccc}
\toprule
Dataset & Setting & Auxiliary &
\multicolumn{2}{c}{$K{=}100$} &
\multicolumn{2}{c}{$K{=}500$} &
\multicolumn{2}{c}{$K{=}1000$} \\
(language) & & &
\multicolumn{2}{c}{CER (\%)} &
\multicolumn{2}{c}{CER (\%)} &
\multicolumn{2}{c}{CER (\%)} \\
& & &
CRNN & HTR-VT~ &
CRNN & HTR-VT~ &
CRNN & HTR-VT~ \\
\midrule

\multirow{4}{*}{\shortstack[c]{KHATT\\\scriptsize(Arabic)}}
& Single-script ($J=0$) & -- 
& 25.95 & 45.46
& 16.68 & 26.03
& 13.25 & 20.21 \\

& Multi-script ($J=1$) & NUST-UHWR
& 20.09 & 26.70
& \textbf{13.53} & 19.96
& \textbf{11.62} & 19.31 \\

& Multi-script ($J=1$) & PHTD
& 23.94 & 33.95
& 15.22 & 22.92
& 12.86 & 19.17 \\

& Multi-script ($J=2$) & NUST-UHWR + PHTD
& \textbf{19.77} & 25.05
& 13.78 & 18.15
& 11.80 & 15.59 \\

\midrule

\multirow{4}{*}{\shortstack[c]{NUST-UHWR\\\scriptsize(Urdu)}}
& Single-script ($J=0$) & -- 
& 26.98 & 52.70
& 13.15 & 25.31
& 10.30 & 19.03 \\

& Multi-script ($J=1$) & KHATT
& 18.37 & 27.35
& 11.37 & 18.68
& 9.48 & 15.61 \\

& Multi-script ($J=1$) & PHTD
& 21.68 & 33.40
& 12.36 & 22.18
& 10.05 & 17.69 \\

& Multi-script ($J=2$) & KHATT + PHTD
& \textbf{17.82} & 25.23
& \textbf{11.24} & 16.66
& \textbf{9.36} & 13.81 \\

\midrule

\multirow{4}{*}{\shortstack[c]{PHTD\\\scriptsize(Persian)}}
& Single-script ($J=0$) & -- 
& 32.50 & 56.75
& 15.81 & 34.04
& 13.40 & 26.78 \\

& Multi-script ($J=1$) & KHATT
& 23.05 & 32.18
& 13.79 & 22.66
& 11.44 & 18.41 \\

& Multi-script ($J=1$) & NUST-UHWR
& 18.59 & 31.27
& \textbf{12.01} & 23.01
& 10.20 & 18.26 \\

& Multi-script ($J=2$) & KHATT + NUST-UHWR
& \textbf{18.27} & 25.45
& 12.31 & 19.52
& \textbf{9.99} & 16.10 \\

\bottomrule
\end{tabular}
}
\end{table*}

\begin{table*}[t]
\caption{Full dataset training ($J=0$ and $J=2$) with $J=0$ comparison to other published methods.}
\label{tab:fullCER}
\centering
\small
\resizebox{0.82\textwidth}{!}{%
\begin{tabular}{ccccc}
\toprule
Dataset & Data & Method & $J{=}0$ & $J{=}2$ \\
(language) & Quantity (K) & & CER (\%) & CER (\%) \\
\midrule

\multirow[c]{3}{*}{\makecell[c]{KHATT\\\scriptsize(Arabic)}} &
\multirow[c]{3}{*}{\makecell[c]{K = 4500}}
& Our CRNN  & 8.80 & 8.59 \\
&& Our HTR-VT & 8.89 & 10.66 \\

\cdashline{3-5}
&& CRNN\cite{al2026cer}, TrOCR\cite{chan2024hatformer}, SRF\cite{saeed2024muharaf}
& ~~8.45, 15.40, 14.10~~ & -- \\

\midrule

\multirow[c]{3}{*}{\makecell[c]{NUST-UHWR\\\scriptsize(Urdu)}} &
\multirow[c]{3}{*}{\makecell[c]{K = 8500}}
& Our CRNN  & 5.48 & 5.50 \\
&& Our HTR-VT & 6.09 & 7.68 \\
\cdashline{3-5}
&& Conv-
transformer\cite{riaz2022conv}, CRNN\cite{al2026cer} & 6.40, 6.66 & -- \\
\midrule

\multirow[c]{3}{*}{\makecell[c]{PHTD\\\scriptsize(Persian)}} &
\multirow[c]{3}{*}{\makecell[c]{K = 1500}}
& Our CRNN  & 12.90 & 9.70 \\
&& Our HTR-VT & 24.70 & 14.80 \\
\cdashline{3-5}
&& CRNN\cite{al2026cer} & 11.30 & -- \\
\bottomrule
\end{tabular}
}
\end{table*}

\subsection{Single-Script Training}

We first analyze single-script training, where the model is trained using only target-language data ($J=0$). As shown in Table~\ref{tab:single_partial_full_across_k}, recognition performance improves as the number of labeled samples increases for both HTR models. 

Across all datasets, the CRNN model performs better than the transformer-based HTR-VT under low-resource settings, with the largest gap observed under the most constrained target-data condition ($K{=}100$), where the transformer model struggles to learn stable and effective recognition. Although HTR-VT improves considerably as the number of labeled samples increases, a noticeable gap compared to CRNN remains across all datasets.
For both architectures, CER decreases steadily from $K=100$ to $K=1000$, indicating that additional target data improves recognition performance. As shown in Table~\ref{tab:fullCER}, this trend generally continues when the full dataset is used in training. Under full data, the performance of both architectures is comparable to or better than previously reported results from the literature.

Performance differences across datasets remain noticeable, particularly at low $K$. To better understand these differences, we analyze the statistical properties of the sampled subsets in Table~\ref{tab:diversity_stats}. As shown in Table~\ref{tab:diversity_stats}, PHTD exhibits substantially lower lexical diversity (e.g., only 738 unique words at $K=1000$), which contributes to its consistently higher CER across both CRNN and HTR-VT compared to KHATT and NUST-UHWR.

\begin{table}[t]
 \caption{Line-level and lexical diversity statistics for the datasets at $K \in \{100, 500, 1000\}$. Duplication ratio is $1 - \text{unique lines}/K$. Mean max Jaccard measures the maximum word overlap between lines. Word TTR is the type–token ratio. Higher duplication and Jaccard, and lower TTR, indicate lower diversity.}
    \label{tab:diversity_stats}
    \centering
    \begin{tabular}{lccccc}
    \midrule
    Dataset & Unique lines & \shortstack[c]{~Duplication~\\ ratio} & \shortstack[c]{~Mean max~\\Jaccard} & ~Unique words~ & Word TTR \\
    \midrule
        KHATT\_K100  & 100  & 0.000 & 0.097 &  859 & 0.735 \\
        KHATT\_K500  & 500  & 0.000 & 0.109 & 3478 & 0.591 \\
        KHATT\_K1000 & 1000 & 0.000 & 0.108 & 6103 & 0.520 \\
        \midrule
        NUST\_K100   & 100  & 0.000 & 0.138 &  688 & 0.597 \\
        NUST\_K500   & 500  & 0.002 & 0.155 & 2356 & 0.410 \\
        NUST\_K1000  & 995  & 0.005 & 0.158 & 3850 & 0.335 \\
        \midrule
        PHTD\_K100   & 93   & 0.070 & 0.665 &  382 & 0.266 \\
        PHTD\_K500   & 389  & 0.222 & 0.782 &  641 & 0.084 \\
        PHTD\_K1000  & 663  & 0.337 & 0.776 &  738 & 0.049 \\
        \midrule
    \end{tabular}
\end{table}

\subsection{Multi-script: One-Auxiliary Training}
We next examine the setting in which the target dataset is combined with a single auxiliary Arabic-script dataset ($J=1$). Across all three languages, one-auxiliary joint training consistently improves recognition compared to single-script training for both CRNN and HTR-VT.


These results suggest that auxiliary dataset size alone does not fully explain the observed gains. Although PHTD is the smallest dataset, it still contributes measurable improvements when used as an auxiliary. Conversely, as a target dataset, PHTD receives the largest relative benefit from cross-lingual training. As reflected in Table~\ref{tab:diversity_stats}, the PHTD training split contains substantial textual overlap and near-duplicate content, reducing its effective diversity. This limited intra-dataset variability likely contributes to the strong gains observed under both one- and two-auxiliary training settings.

These observations suggest that transfer effectiveness depends not only on dataset size, but also on the degree of internal diversity within the target dataset. Cross-lingual training appears most beneficial when the target corpus is constrained in size and limited in diversity.

Overall, CRNN consistently outperforms HTR-VT across low-resource settings, particularly under the most limited target-data condition ($K{=}100$), where the transformer model struggles to learn effective recognition from very limited training data.

\subsection{Multi-script: Two-Auxiliary Joint Training}

We next extend the analysis to the setting in which the target dataset is combined with both remaining Arabic-script datasets ($J=2$). Across all three languages, two-auxiliary joint training consistently improves performance over single-script training for both CRNN and HTR-VT.

For the CRNN model, KHATT CER decreases from    25.95\% to 19.77\% at $K=100$, and from 16.68\% to 13.78\% at $K=500$. Gains remain positive at $K=1000$. 
For NUST-UHWR, two-auxiliary training  reduces CER from 26.98\% to 17.82\% at $K=100$, and from 13.15\% to 11.24\% at $K=500$. At $K=1000$, improvements remain but are smaller.
For PHTD, consistent gains are observed across all regimes: CER decreases from 32.50\% to 18.27\% at $K=100$, from 15.81\% to 12.31\% at $K=500$, and from 13.40\% to 9.99\% at $K=1000$.

Similar patterns are observed for HTR-VT, although the transformer model remains less data-efficient than CRNN under limited target-language data. 
Two-auxiliary joint training consistently improves recognition compared to single-script training across all datasets when training data is scarce. For datasets with larger amounts of training data (KHATT and NUST-UHWR), the transformer exhibits better performance with the homogeneous data of a single source ($J=0$). As shown in Table~\ref{tab:fullCER}, the benefits of cross-language joint training become less consistent for HTR-VT in full-data settings, whereas CRNN continues to exhibit stable gains or comparable performance.

Overall, the use of two auxiliary datasets provides the most consistent improvements under low-resource conditions, particularly when the target dataset is both small and limited in diversity.

When trained on the full target datasets, Table~\ref{tab:fullCER}, the performance gap between CRNN and HTR-VT becomes much smaller for KHATT and NUST-UHWR. For example, on NUST-UHWR the HTR-VT achieves a CER of 6.09\% compared to 5.48\% for CRNN, while on KHATT the HTR-VT reaches 8.89\% compared to 8.80\% for CRNN.  In contrast, a larger gap remains on PHTD, whose full training set contains only approximately 1500 lines, where HTR-VT achieves 24.70\% CER compared to 12.90\% for CRNN, but with the assistance of KHATT and NUST-UHWR ($J=2$) PHTD can achieve 9.7\% CER with the CRNN and 14.80\% under HTR-VT. 


While we and \cite{al2026cer} both use a CRNN, minor differences between our CRNN results and \cite{al2026cer} are likely attributable to differences in preprocessing and resizing strategy.
In our experiments, a unified resizing configuration derived from the average dimensions across all three datasets was used to enable consistent cross-lingual training, whereas \cite{al2026cer} used dataset-specific resizing configurations computed separately for each dataset.

\begin{figure}[t]
\centering
\includegraphics[width=0.75\textwidth]{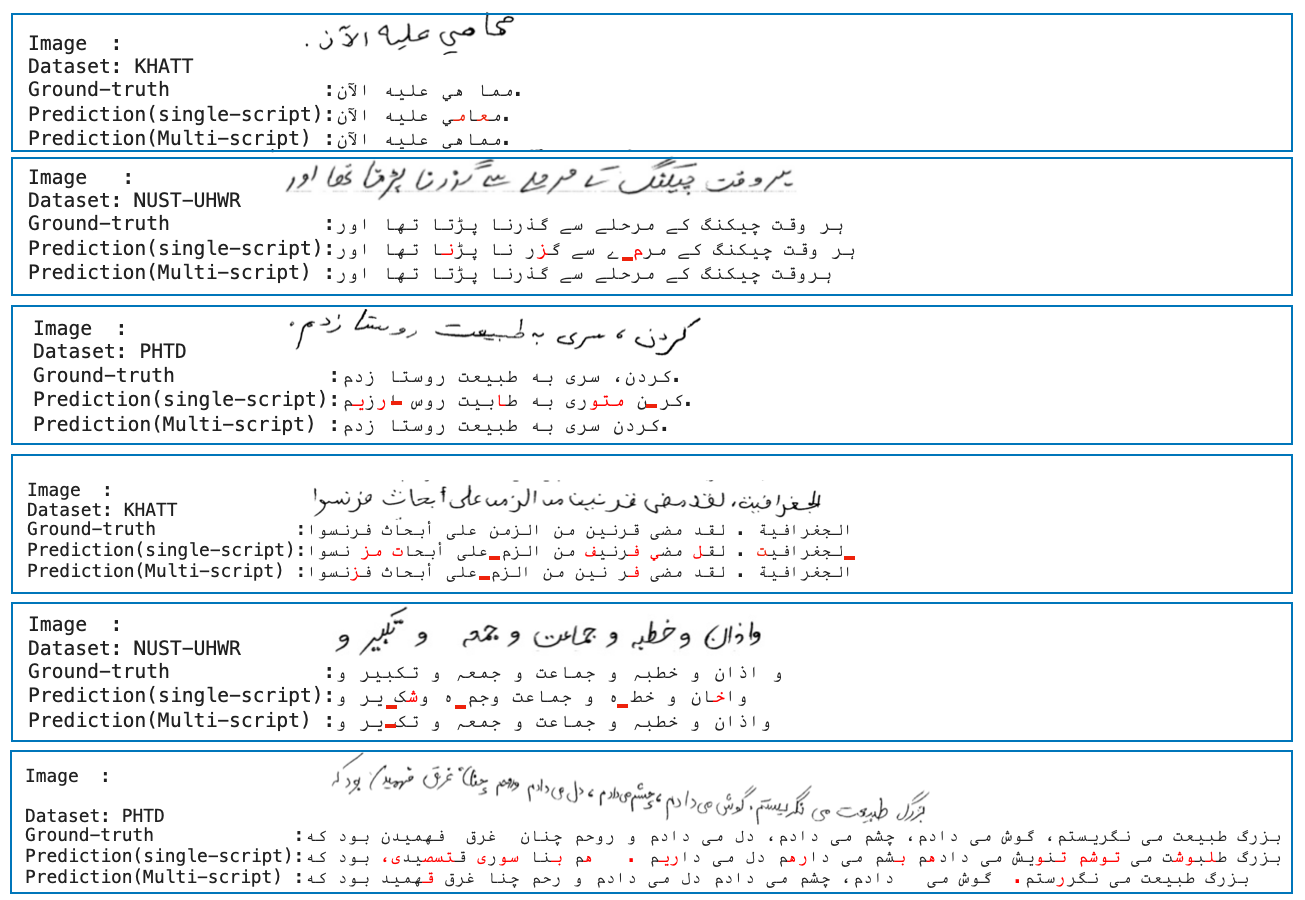}
\caption {Qualitative comparison between single-script and multi-script CRNN training at $K{=}1000$. For each dataset (KHATT, NUST-UHWR, PHTD), two representative examples are shown. Predictions from single-script training ($J=0$) are contrasted with those from multi-script ($J=2$) joint training. Characters incorrectly predicted are highlighted in red.} \label{fig:qualitative}
\end{figure}

Fig.~\ref{fig:qualitative} provides representative examples demonstrating how cross-lingual joint training reduces character-level errors compared to single-script training with the CRNN.

\subsection{Evaluation  on an additional low-resource dataset}

The results on KHATT, NUST-UHWR, and PHTD demonstrate that cross-lingual joint training consistently improves recognition under low-resource settings. To assess whether this effect generalizes beyond the primary experimental setting, we evaluate the proposed method on another Urdu handwritten dataset, UNHD. Since the previous experiments showed that CRNN models remain more effective under extremely limited target-data conditions, we focus this additional evaluation on the CRNN architecture.

UNHD contains 10,000 text lines written by 500 writers; however, only approximately 600 training lines are unique in terms of semantic content, resulting in limited effective lexical diversity \cite{anjum2023caltext}. This characteristic makes UNHD particularly suitable for evaluating robustness under  real-world  low-resource conditions and low diversity.

We follow the same experimental protocol used for the primary datasets. The CRNN architecture, optimization strategy, preprocessing, and hyperparameters remain unchanged. For multi-script training, we use Arabic (KHATT) and Persian (PHTD) as auxiliary datasets. We exclude NUST-UHWR to avoid same-language transfer, ensuring that any improvements arise strictly from cross-lingual information.  As shown in Table~\ref{tab:unhd_sota_with_iam}, the single-script achieves performance comparable to the previously reported state-of-the-art obtained using the full UNHD training set. In contrast, multi-script joint training reduces CER from 17.20\% to 14.45\%, achieving a substantial improvement over the single-script baseline under the same setting without introducing additional Urdu data.

Overall, cross-lingual joint training yields consistent improvements across
datasets, with the largest gains observed when the target data is both limited in size and low in diversity.

\begin{table}[t]
\centering
\caption{CRNN performance  on the UNHD ($K{=}600$, full target set).}
\label{tab:unhd_sota_with_iam}
\small
\setlength{\tabcolsep}{6pt}
\resizebox{0.85\textwidth}{!}{%
\begin{tabular}{l c c c}
\toprule
Dataset (language) 
& Setting 
& Auxiliary 
& CER (\%) \\
\midrule

\multirow{5}{*}{\shortstack[l]{UNHD \cite{ahmed2019handwritten}\\\scriptsize(Urdu)}}

& Single-script ($J=0$)
& -- 
& 17.20 $\pm$ 0.02  \\

\cmidrule(lr){2-4}

& Multi-script ($J=1$)
& KHATT {\scriptsize(Arabic)}
& 14.79 $\pm$ 0.07 \\

\cmidrule(lr){2-4}

& Multi-script ($J=1$)
& PHTD {\scriptsize (Persian)} 
& 15.60 $\pm$ 0.32  \\

\cmidrule(lr){2-4}

& Multi-script ($J=2$)
& KHATT {\scriptsize(Arabic)} + PHTD {\scriptsize (Persian)} 
& \textbf{14.45} $\pm$ 0.13 \\



\bottomrule
\end{tabular}
}

\end{table}

\subsection{Character-Level Transfer: Shared vs. Unique Characters}
To quantify character-level transfer for the CRNN model, we compare single-script training 
($J=0$) and multi-script training with two auxiliary datasets ($J=2$) 
at $K=1000$.  For each character $c$, we compute

\begin{equation}
\Delta \mathrm{CER}(c)
=
\mathrm{CER}_{\mathrm{Multi}}(c)
-
\mathrm{CER}_{\mathrm{Single}}(c).
\end{equation}
Negative values of $\Delta \mathrm{CER}(c)$ indicate desirable reduced recognition error 
under multi-script joint training.
\newline
\newline
\noindent\textit{Quantifying character-level transfer.}
We quantify transfer effects by computing the mean per-character recognition error change separately for characters shared across scripts and language-specific (unique) characters (Table~\ref{tab:shared_unique_transfer}). On Urdu (NUST-UHWR), shared characters improve on average ($-1.97$ pp), whereas unique characters degrade ($+2.02$ pp), with a statistically significant difference (Welch $p=0.0021$). This indicates that cross-lingual joint training primarily benefits shared characters, while language-specific symbols may experience mild interference under joint optimization.

\begin{figure}[t]
\centering
\includegraphics[width=0.95\textwidth]{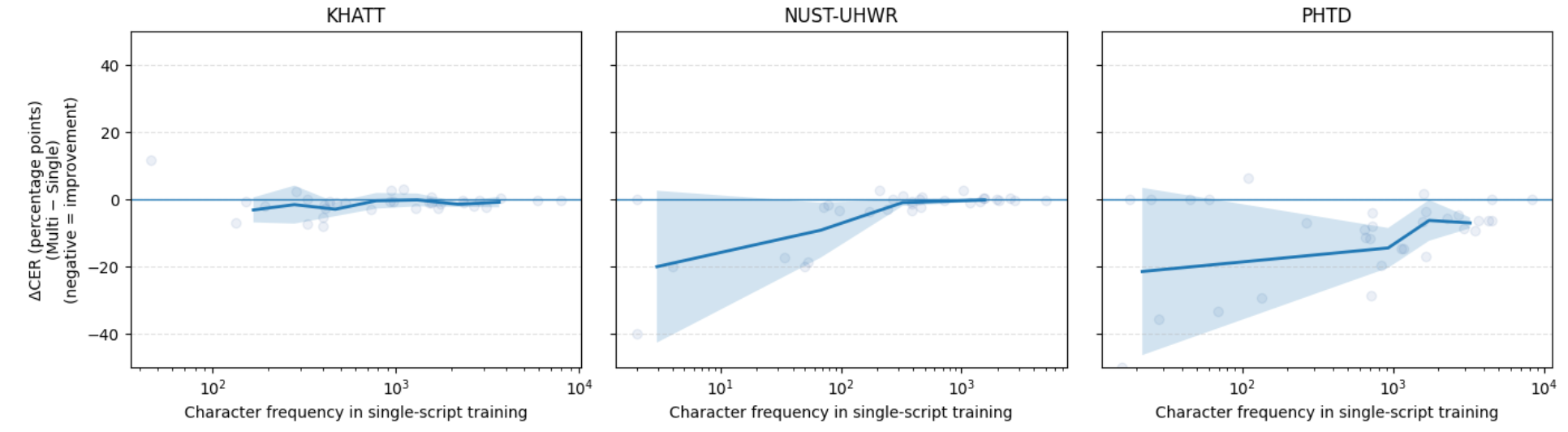}
\caption{CRNN Character-level transfer for shared characters. 
Each point represents a character; the $x$-axis shows its frequency (log scale), and the $y$-axis shows $\Delta \mathrm{CER}(c)=\mathrm{CER}_{\text{Multi}}(c)-\mathrm{CER}_{\text{Single}}(c)$ (negative values indicate improvement). 
The solid line shows the mean $\Delta \mathrm{CER}$ across frequency ranges.}
\label{fig:shared_all_binned}
\end{figure}

For Arabic (KHATT), improvements are modest and similar for shared and unique characters, though only one unique character is present. For Persian (PHTD), which contains no unique characters under our grouping, substantial improvements are observed across shared symbols ($-11.21$ pp on average), consistent with its smaller dataset size and stronger reliance on auxiliary data.
\begin{table}[t]
\caption{CRNN Mean per-character recognition error change 
($\Delta \mathrm{CER} = \mathrm{CER}_{\text{Multi}} - \mathrm{CER}_{\text{Single}}$, in percentage points). 
Negative values indicate improvement  (reducing CER) under multi-script training. 
$p$-value computed using Welch’s $t$-test comparing shared and unique characters.}
\label{tab:shared_unique_transfer}
\centering
\small
 \resizebox{0.7\textwidth}{!}{
\begin{tabular}{lccccc}
\toprule
Dataset & $n_{\text{shared}}$ & $n_{\text{unique}}$ &
Mean $\Delta$ (shared) & Mean $\Delta$ (unique) & $p$ \\
\midrule
KHATT & 35 & 1 & $-0.94$ & $-0.79$ & -- \\
NUST-UHWR  & 33 & 7 & $-1.97$ & $+2.02$ & 0.002 \\
PHTD  & 38 & 0 & $-11.21$ & -- & -- \\
\bottomrule
\end{tabular}
}
\end{table}
\newline
\newline
\noindent\textit{Shared characters.}
Fig.~\ref{fig:shared_all_binned} further analyzes shared characters by plotting $\Delta\mathrm{CER}$ as a function of single-script character frequency (log scale). Each point corresponds to one shared character. Across KHATT, NUST-UHWR, and PHTD, \textbf{larger gains are observed for low-frequency shared characters}, with improvements decreasing as target-side frequency increases. This indicates that auxiliary data primarily compensates for target-side data sparsity rather than uniformly improving all characters.

Table~\ref{tab:subtables_shared} lists the shared characters exhibiting the largest error reductions for each dataset (single-script count $\ge 15$). Improvements are most pronounced for rare and visually confusable letters (e.g., \ar{ث}, \ar{ض}, \ar{ظ}), while high-frequency characters show comparatively minor changes, suggesting a diminishing-return effect for already well-learned symbols.
\begin{table*}[t]
\centering
\caption{Shared characters with the largest CER reductions under cross-lingual training. 
Negative $\Delta \mathrm{CER} $ values indicate improvement.}

\label{tab:subtables_shared}
\small
\setlength{\tabcolsep}{1pt}
\renewcommand{\arraystretch}{1.05}
\begin{subtable}[b]{0.30\textwidth}
\caption{\scriptsize KHATT (Arabic)}
\centering
\resizebox{\textwidth}{!}{%
\begin{tabular}{c cc cc c}
\toprule
 & \multicolumn{2}{c}{Train} & \multicolumn{2}{c}{CER (\%)} &  \\
\cmidrule(lr){2-3}\cmidrule(lr){4-5}
Char & Single & Multi & Single & Multi & $\Delta$ (pp) \\
\midrule
\ar{ث} & 399  & 891   & 30.5 & 22.5 & $-8.1$ \\
\ar{ض} & 326  & 1365  & 24.5 & 17.4 & $-7.1$ \\
\ar{ظ} & 134  & 872   & 22.2 & 15.4 & $-6.8$ \\
\ar{خ} & 398  & 3705  & 26.8 & 21.5 & $-5.3$ \\
\ar{ج} & 730  & 7492  & 16.5 & 13.5 & $-2.9$ \\
\bottomrule
\end{tabular}}
\end{subtable}%
\hspace{0.03\textwidth}%
\begin{subtable}[b]{0.30\textwidth}
\caption{\scriptsize NUST-UHWR (Urdu)}
\centering
\resizebox{\textwidth}{!}{%
\begin{tabular}{c cc cc c}
\toprule
 & \multicolumn{2}{c}{Train} & \multicolumn{2}{c}{CER (\%)} &  \\
\cmidrule(lr){2-3}\cmidrule(lr){4-5}
Char & Single & Multi & Single & Multi & $\Delta$ (pp) \\
\midrule
\ar{ذ} & 50   & 2054  & 47.3 & 27.3 & $-20.0$ \\
\ar{ث} & 53   & 1901  & 58.1 & 39.5 & $-18.6$ \\
\ar{ء} & 34   & 849   & 34.5 & 17.2 & $-17.2$ \\
\ar{ط} & 174  & 3173  & 11.0 & 7.5  & $-3.5$ \\
\ar{ض} & 97   & 1857  & 20.5 & 17.2 & $-3.3$ \\
\bottomrule
\end{tabular}}
\end{subtable}%
\hspace{0.02\textwidth}%
\begin{subtable}[b]{0.30\textwidth}
\caption{\scriptsize PHTD (Persian)}
\centering
\resizebox{\textwidth}{!}{%
\begin{tabular}{c cc cc c}
\toprule
 & \multicolumn{2}{c}{Train} & \multicolumn{2}{c}{CER (\%)} &  \\
\cmidrule(lr){2-3}\cmidrule(lr){4-5}
Char & Single & Multi & Single & Multi & $\Delta$ (pp) \\
\midrule
\ar{ء} & 16   & 1043  & 50.0 & 0.0  & $-50.0$ \\
\ar{ذ} & 28   & 2544  & 71.4 & 35.7 & $-35.7$ \\
\ar{ظ} & 69   & 1324  & 60.0 & 26.7 & $-33.3$ \\
\ar{غ} & 135  & 2109  & 70.8 & 41.7 & $-29.2$ \\
\ar{ج} & 711  & 9904  & 40.5 & 11.9 & $-28.6$ \\
\bottomrule
\end{tabular}}
\end{subtable}
\end{table*}
\begin{table*}[t]
\caption{Unique characters under cross-lingual training. 
$\Delta \mathrm{CER}$ values close to zero indicate limited transfer to language-specific symbols.}

\label{tab:unique_chars_cer}
\centering
\small
\setlength{\tabcolsep}{3pt}
\renewcommand{\arraystretch}{1.05}
\begin{subtable}[t]{0.4\textwidth}
\caption{\scriptsize KHATT (Arabic)}
\centering
\resizebox{0.95\textwidth}{!}{%
\begin{tabular}{c cc cc c}
\toprule
 & \multicolumn{2}{c}{Train} & \multicolumn{2}{c}{CER (\%)} &  \\
\cmidrule(lr){2-3}\cmidrule(lr){4-5}
Char & Single & Multi & Single & Multi & $\Delta CER$ (pp) \\
\midrule
\ar{إ} & 412 & 412 & 12.07 & 11.29 & -0.79 \\
\bottomrule
\end{tabular}}
\end{subtable}%
\hfill
\begin{subtable}[t]{0.4\textwidth}
\caption{\scriptsize NUST-UHWR (Urdu)}
\centering
\resizebox{0.95\textwidth}{!}{%
\begin{tabular}{c cc cc c}
\toprule
 & \multicolumn{2}{c}{Train} & \multicolumn{2}{c}{CER (\%)} &  \\
\cmidrule(lr){2-3}\cmidrule(lr){4-5}
Char & Single & Multi & Single & Multi & $\Delta CER$ (pp) \\
\midrule
\ar{ۂ} & 2    & 2    & 100.00 & 100.00 & 0.00 \\
\ar{ھ} & 570  & 570  & 9.99   & 10.13  & 0.15 \\
\ar{ے} & 2097 & 2097 & 1.64   & 2.25   & 0.61 \\
\ar{ہ} & 2214 & 2214 & 9.01   & 9.76   & 0.75 \\
\ar{ں} & 980  & 980  & 3.74   & 4.52   & 0.78 \\
\ar{ڈ} & 117  & 117  & 13.18  & 17.05  & 3.88 \\
\ar{ٹ} & 303  & 303  & 21.75  & 25.71  & 3.95 \\
\ar{ڑ} & 114  & 114  & 17.74  & 21.77  & 4.03 \\
\bottomrule
\end{tabular}}
\end{subtable}
\end{table*}
\newline
\newline
\noindent\textit{Unique characters.}
As shown in Fig.\ref{fig:venn}, not all letters are shared across languages. we examine characters unique to each dataset, which receive no additional exposure from auxiliary languages. Results are shown in Table~\ref{tab:unique_chars_cer}. For KHATT, the single unique character (\ar{إ}) remains largely stable ($-0.79$ pp). In NUST-UHWR, most Urdu-specific characters exhibit small fluctuations (typically within $\pm 0.8$ pp), consistent with training variability. However, a subset (\ar{ٹ}, \ar{ڈ}, \ar{ڑ}) shows moderate increases in error ($+3$ to $+4$ pp), indicating competitive capacity allocation in favor of shared characters during joint training.

Overall, these findings show that cross-lingual joint training primarily benefits shared characters, particularly those that are infrequent in the target data, while unique (language-specific) characters remain stable or may experience limited degradation. These results support the hypothesis that transfer effects are primarily driven by script-level overlap rather than uniform improvements across the character inventory.


\section{Conclusion}\label{sec:conclusion}

This work presents a controlled line-level evaluation of cross-lingual joint training for Arabic-script HTR under low-resource conditions, using both CRNN and transformer-based HTR-VT models trained across Arabic, Urdu, and Persian datasets while systematically restricting the amount of labeled target-language data. We isolate and quantify the effect of cross-lingual transfer. Across datasets, joint training consistently improves recognition over single-script baselines when labeled target data is limited.

While both HTR architectures benefit from cross-lingual joint training under low-resource conditions, the experiments show that CRNN models remain more data-efficient under extremely low-resource conditions. In contrast, the transformer-based HTR-VT model struggles to learn effective recognition at very small target sizes ($K{=}100$), although its performance improves as more target-language data becomes available and it benefits most from cross-lingual training under low-resource conditions.

The largest gains are observed for datasets with limited size and diversity, particularly PHTD and UNHD. Character-level analysis further shows that improvements are concentrated on shared characters, especially infrequent ones, while language-specific characters receive limited benefit. These findings indicate that transfer effectiveness depends on both character overlap and target-dataset diversity.

Overall, these results demonstrate that related Arabic-script datasets can effectively complement one another under data scarcity. Cross-lingual joint training provides a practical strategy for improving recognition performance without collecting additional target-language annotations or relying on large-scale pretraining. Furthermore, under single-script training ($J{=}0$) using the full target datasets, the performance gap between CRNN and HTR-VT becomes relatively small for KHATT and NUST-UHWR, suggesting that transformer-based architectures become increasingly competitive as sufficient target-language data becomes available.

suggesting that transformer-based architectures can become increasingly competitive as sufficient target-language data becomes available.

Future work will investigate how transfer dynamics evolve as target-language data increases and will compare joint training with alternative paradigms such as pretraining followed by fine-tuning.


%
\bibliography{ref}
\bibliographystyle{elsarticle-num}





\end{document}